\documentclass{article} 
\usepackage{iclr2025_conference,times}

\usepackage{amsmath,amsfonts,bm}

\def\eqref#1{equation~\ref{#1}}

\def\1{\bm{1}}

\DeclareMathAlphabet{\mathsfit}{\encodingdefault}{\sfdefault}{m}{sl}
\SetMathAlphabet{\mathsfit}{bold}{\encodingdefault}{\sfdefault}{bx}{n}

\usepackage[hidelinks,breaklinks=true,colorlinks,bookmarks=false,citecolor=citecolor,linkcolor=linkcolor]{hyperref}
\usepackage{url}
\usepackage[utf8]{inputenc} 
\usepackage[T1]{fontenc}    
\usepackage{url}            
\usepackage{booktabs}       
\usepackage{amsfonts}       
\usepackage{nicefrac}       
\usepackage{microtype}      
\usepackage{array}
\newcolumntype{C}[1]{>{\centering\arraybackslash}m{#1}}
\usepackage{xcolor}         
\usepackage{color, colortbl}
\usepackage{xspace}
\usepackage[shortlabels]{enumitem}
\usepackage{boxedminipage}
\usepackage{wrapfig}

\definecolor{citecolor}{HTML}{2980b9}
\definecolor{linkcolor}{HTML}{c0392b}
\definecolor{darkorange}{HTML}{FF8C00}
\definecolor{chocolate}{HTML}{D2691E}
\definecolor{darkgreen}{HTML}{006400}
\definecolor{darkblue}{HTML}{00008B}
\definecolor{mediumblue}{HTML}{0000CD}
\definecolor{dodgerblue}{HTML}{1E90FF}
\definecolor{royalblue}{HTML}{4169E1}
\definecolor{shadecolor}{RGB}{237,237,237}
\definecolor{backred}{RGB}{255, 190, 190}
\definecolor{backblue}{RGB}{210, 230, 250}

\definecolor{zrrgreen}{HTML}{008000}
\definecolor{zrrblue}{HTML}{4682B4}
\definecolor{zrrred}{HTML}{B22222}

\usepackage{amsmath}
\usepackage{amssymb}
\usepackage{graphicx}

\usepackage{multirow}
\usepackage{makecell}
\usepackage{caption}

\usepackage{adjustbox}
\usepackage{wrapfig}

\usepackage{float}
\usepackage{subfig}

\newcommand{\name}{\textsc{ProbleMathic}\xspace}

\definecolor{light_green}{HTML}{b2ffb2}
\definecolor{sage}{HTML}{c3efb2}
\definecolor{light_red}{HTML}{ffb2b2}
\definecolor{light_blue}{HTML}{add8e6}

\usepackage[most]{tcolorbox}
\tcbset{on line, 
        boxsep=1pt, left=0pt,right=0pt,top=0pt,bottom=0pt,
        colframe=white,
        colback=light_green,  
        highlight math style={enhanced}
        }

\title{Cutting Through the Noise: Boosting LLM\\Performance on Math Word Problems}

\iclrfinalcopy 
\begin{document}

\author
{Ujjwala Anantheswaran $^{1\diamondsuit*}$ \quad Himanshu Gupta $^{1\ddagger*}$ \quad  Kevin Scaria $^{1\ddagger*}$ \quad \textbf{Shreyas Verma}$^{2}$\quad\\  \textbf{Chitta Baral}$^{1}$ \quad \textbf{Swaroop Mishra}$^{1\dagger}$  \\
\small{$^{1}$Arizona State University} \quad
\small{$^{2}$Georgia Institute of Technology} \quad \\
\tt\small {\{uananthe, hgupta35, kscaria\}}@asu.edu
}
\maketitle

\begin{abstract}
Large Language Models (LLMs) excel at various tasks, including solving math word problems (MWPs), but struggle with real-world problems containing irrelevant information. 
To address this, we propose a prompting framework that generates adversarial variants of MWPs by adding irrelevant variables. 
We introduce a dataset, \name containing both adversarial and non-adversarial MWPs.
Our experiments reveal that LLMs are susceptible to distraction by numerical noise, resulting in an average relative performance drop of $\sim26\%$ on adversarial MWPs.
To mitigate this, we fine-tune LLMs (Qwen-2, Mistral) on the adversarial samples from our dataset. 
Fine-tuning on adversarial training instances improves performance on adversarial MWPs by $\sim8\%$, indicating increased robustness to noise and improved ability to identify relevant data for reasoning. 
Finally, to assess the generalizability of our prompting framework, we introduce GSM-8K-Adv, an adversarial variant of the GSM-8K benchmark. 
LLMs continue to struggle when faced with adversarial information, reducing performance by up to $24\%$. 
\footnote{$\diamondsuit$ Currently in Microsoft \quad $*$ Equal Contribution\\ $\ddagger$ Currently in Amazon (The work was done prior to joining Amazon) \quad $\dagger$ Currently in Google DeepMind \\ Codebase: \url{https://github.com/him1411/problemathic} \\ Dataset: \url{https://huggingface.co/datasets/him1411/problemathic} }.
\end{abstract}

\section{Introduction}

\begin{wrapfigure}{r}{0.45\textwidth}
\centering
 \vspace{-4.5mm}
	\includegraphics[height= 9 cm]{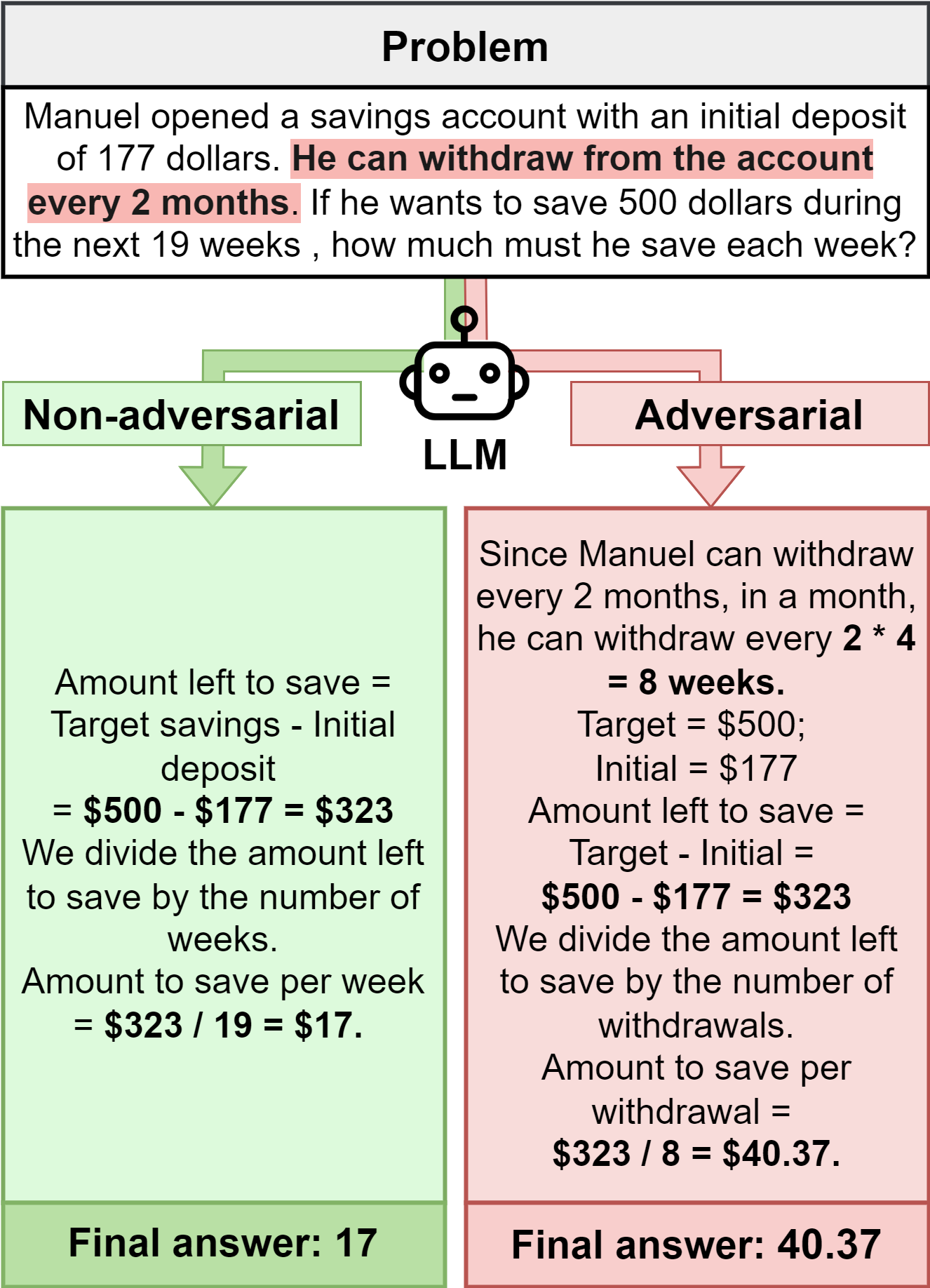}
	\caption{An example showing how adding noise (highlighted in $\tcbox[colback=light_red]{red}$) to an MWP results in incorrect reasoning and an inaccurate solution. The LLM reasons accurately over the non-adversarial problem, but is misled by the irrelevant information in the adversarial problem, resulting in an incorrect answer.  
 }
\label{teaser}
\end{wrapfigure}

Large Language Models (LLMs) \citep{llama3modelcard,team2023gemini,achiam2023gpt} have achieved impressive performance on various tasks, including mathematical problem solving \citep{imani-etal-2023-mathprompter,gaur-saunshi-2023-reasoning,RomeraParedes2023MathematicalDF,ahn2024large}. 
Existing mathematical datasets \citep{patel-etal-2021-nlp,Sawada2023ARBAR,mishra-etal-2022-lila,gupta2023john} typically feature simplified questions with limited variables and numerical data directly relevant to the problem. 
However, real-world MWPs arise within the context of broader reasoning tasks. Thus, they are likely to contain irrelevant information stemming from the wider text. 
This unrelated information acts as noise, distracting the reasoning process of language models (Fig.~\ref{teaser}). 
Recent works have shown that LLMs do not perform consistently well on mathematical reasoning \citep{ahn2024large,shakarian2023independent,Chang2023ASO} and are sensitive to linguistic variations in MWPs \citep{patel-etal-2021-nlp,Shi2023LargeLM,kumar-etal-2021-adversarial-examples}. 
Previous works \citep{kumar-etal-2022-practice} have attempted to make LLMs more robust by generating "adversarial" training data but fail to emulate real-world data by not introducing textual noise.

To initiate a systematic study, we introduce a new dataset \name containing both adversarial and non-adversarial mathematical word problem (MWP) pairs. 
The dataset includes problems of two difficulty levels: \textit{Simple} and \textit{Complex}.
These adversarial MWPs are augmented to add irrelevant numerical information while maintaining the integrity of the original question; i.e. their solution and final answer are the same as their non-adversarial counterparts. 
We employ various LLMs, including Gemini-1.5 Pro, Claude-3.5 Sonnet, and Qwen-1.5, among others, for inference. 
Our observations reveal significant performance drops (16.07\%, 22.93\%, 41.53\%, respectively) between adversarial and non-adversarial problems, highlighting the models' inability to reason accurately with noisy data. 
To address this challenge, we propose a prompting framework that allows us to leverage existing MWPs of varying complexities and generate augmented adversarial versions for the training set. 
We demonstrate that fine-tuning LLMs on these samples enhances their ability to identify irrelevant variables. 
Fine-tuning Llama-2 (7B, 13B) and Mistral (7B) on these samples yields an average 8\% improvement in performance on both Simple (8.19\%) and Complex (7.07\%) adversarial test samples.
Finally, to assess the generalizability of our prompting framework, we create an adversarial variant of the widely used GSM-8K benchmark (GSM-8K-Adv). 
This reveals that LLMs (Mistral Large, Llama-2, and Llama-3) experience an average performance drop of $\sim4\%$ when tested on this adversarial benchmark. 
Additionally, we compare our approach with others in the domain of adversarial data generation in \S \ref{sec:comparison}.

\noindent\textbf{Contributions:}
(a) We introduce \name, a dataset of \textit{Simple} and \textit{Complex} MWPs that demonstrates the susceptibility of LLMs to irrelevant numerical information. (b) We propose a prompting framework to generate adversarial variants of existing MWPs. We show that fine-tuning on these adversarial samples leads to improved performance and robustness to noise. 
(c) Using this prompting framework, we introduce GSM-8K-Adv, an adversarial variant of the GSM-8K benchmark.

\section{Methodology}

\subsection{Adversarial data generation}
\label{sec:adversarial_data_generation}

We propose a constrained additive approach, wherein we add numerical information (\textit{Structural Invariance}) to the original MWP. 
We use our hand-crafted set of MWPs as seed samples and augment them in accordance with the following constraints:
(1) The added variables must not be related to or derived from the existing variables in the passage. (2) The added variables must not share the same physical unit as any of the original variables. (3) The augmented text should not add any numerical information about existing variables that did not exist in the original passage. 
This gives us a set of adversarial MWPs containing variables that do not feature in their solution equations.
Hence, this allows us to evaluate the model's ability to identify relevant information and demonstrate reasoning ability beyond logically connecting all numerical entities in the input in order to produce a solution.
The prompting framework used to adversarially augment the seed samples can be found in \S \ref{prompts}. 
We also compare our adversarial data generation with other contemporaneous approaches (\S \ref{sec:comparison}).

\begin{table*}[t!]
\resizebox{\linewidth}{!}
{
    \begin{tabular}{l|l}
\toprule
\textbf{Sample Type} &
  \textbf{Sample} \\ \midrule
\begin{tabular}[c]{@{}l@{}}\textbf{\textit{Simple}} \\ \textbf{Adv}\\ \textbf{Train} \\ 1313\end{tabular} &
  \begin{tabular}[c]{@{}l@{}}A pet supply store has 600 bags of dog food \textcolor{red}{\textbf{weighing 50 kg}} and 327 bags of cat food \textcolor{red}{\textbf{bringing the total weight to 100 kg.}} \\ How many more bags of dog food are there than cat food?\\ \\
  \textcolor{blue}{\textbf{Solution:} The weight of the bags is not relevant to the question. The question asks about the difference in the number of }\\ \textcolor{blue}{bags of dog food and cat food.}\\ \textcolor{blue}{\textbf{Relevant variables}: Number of bags of dog food and cat food (600 and 327)} \\ \textcolor{blue}{\textbf{Explanation}: The store has 600 bags of dog food and 327 bags of cat food. To find out how many more bags of dog food }\\ \textcolor{blue}{there are, we subtract the number of cat food bags from the number of dog food bags. So, 600 - 327 = 273. Thus, there }\\ \textcolor{blue}{are 273 more bags of dog food than cat food.} \\ \textbf{Answer: 273}\end{tabular} \\
 &
   \\
\begin{tabular}[c]{@{}l@{}}\textbf{\textit{Complex}} \\ \textbf{Adv}\\ \textbf{Train}\\ 613\end{tabular} &
  \begin{tabular}[c]{@{}l@{}}A pet shelter had 8 puppies \textcolor{red}{\textbf{living in 500 cubic feet}} when another 19 were brought in. \textcolor{red}{\textbf{The volume of the shelter increased}} \\ \textcolor{red}{\textbf{to 700 cubic feet}}. If 3 puppies a day are adopted, how long would it take for all of them to be adopted? \\ \\
  \textcolor{blue}{\textbf{Solution}: As the problem deals with the number of days it would take for all puppies to be adopted, we need to consider} \\ \textcolor{blue}{the total number of puppies and the rate at which they are adopted. The volume of the shelter is irrelevant to this calculation.}\\ \textcolor{blue}{\textbf{Relevant variables}: Number of puppies (Changes from 8 to 27), Number of puppies adopted per day (Constant at 3)}\\ \textcolor{blue}{\textbf{Solution}: The pet shelter initially had 8 puppies and then 19 more were brought in, making a total of 8 + 19 = 27 puppies.} \\ \textcolor{blue}{If 3 puppies are adopted each day, it would take 27 / 3 = 9 days for all of them to be adopted.}\\ \textbf{Answer: 9.0} \\ \end{tabular} 
   \\ \midrule
\begin{tabular}[c]{@{}l@{}}\textit{\textbf{Simple}} \\ \textbf{Adv}\\ \textbf{Test}\\ 60\end{tabular} &
  \begin{tabular}[c]{@{}l@{}}In the drawer, there are 11 rulers \textcolor{red}{\textbf{and 34 crayons. The drawer is 80 percent full.}} Tim added 14 larger rulers to the drawer. \\ \textcolor{red}{\textbf{The drawer becomes 95 percent full.}} How many rulers are now there in all ?  \\ \textbf{Answer: 25}\end{tabular} \\
 &
   \\
\begin{tabular}[c]{@{}l@{}}\textit{\textbf{Complex}} \\ \textbf{Adv}\\ \textbf{Test}\\ 73\end{tabular} &
  \begin{tabular}[c]{@{}l@{}}Dave, who was helping the cafeteria workers, could only carry 9 trays at a time. \textcolor{red}{\textbf{He started with a distance of 50 meters to}} \\ \textcolor{red}{\textbf{cover, and he completed it in 20 seconds.}} If he had to pick up 17 trays from one table and 55 trays from another, how many\\ trips will he make?\\ \textbf{Answer: 8}\end{tabular} \\ \bottomrule
\end{tabular}
}
\caption{Examples of generated adversarial train instances and hand-crafted adversarial test instances for \textit{Simple} and \textit{Complex} samples. The adversarial augmentations are highlighted in \textbf{\textcolor{red}{red}}. We fine-tune models to identify irrelevant variables and reason over relevant ones as shown in the given \textcolor{blue}{\textbf{solutions}}.}
\label{tab:datastats}
\end{table*}

\subsection{Dataset}
\begin{table}[]
\centering
\resizebox{\textwidth}{!}
{
\begin{tabular}{p{0.04\textwidth}|p{0.4\textwidth}|p{0.4\textwidth}|p{0.1\textwidth}|p{0.1\textwidth}}
 \toprule
 \textbf{Type}& \textbf{Sample}& \textbf{Explanation}& \textbf{ Samples (\textit{Simp.})} & \textbf{ Samples (\textit{Compl.})} \\ \midrule
 A &  Louis ate 54 Lemon Heads,  which came in packages of 6.\textbf{ Lemon heads have a 3\% acidity rating. The maximum acidity  consumption for a human should not  exceed 20\%}. How many whole boxes did he eat?&  Lemon heads are a relevant entity in the context of this problem. However, acidity level is not a variable that has any relation with the question regarding the number of Lemon heads and boxes.& 22& 20\\ \midrule
 B  &  While making brownies for a bake sale,  Victor used 0.625 of a scoop of brown  sugar as well as 0.25 of a scoop of white sugar. \textbf{To make cookies,  he used 0.875 of a scoop of brown sugar and 0.5 of a scoop of  white sugar.} How much more brown  sugar did Victor use for brownies?&  While the added variables (weight of brown sugar and white sugar) are  directly relevant to the problem,  their association with the entity  “cookies”, which is irrelevant to the question being asked, renders this added numerical information irrelevant.& 8& 10\\ \midrule
 C  &  A toy store had 6 giant stuffed bears in stock. After receiving a new shipment, they now have 24 bears. Each bear weighs a pound. The store puts these bears onto shelves with 6 on each shelf. \textbf{The shelf capacity is only 3 pounds, but it holds up to 10 pounds without breaking.} How many shelves did they use? &  The shelf is an entity that is relevant in the context of this problem. Furthermore, shelf capacity is relevant to the problem; however, given its range, does not have an effect on the solution. The variable value thus makes the addition irrelevant to the solution.  & 12 & 8 \\ \midrule
 D  &  Oliver had 35 dollars in his wallet. On his birthday, he got 50 more dollars as a gift. He was very excited and  decided to buy a new game that he had been wanting for a long time,\textbf{ along with  some apples that cost 3 pounds per  kilogram.} The game cost him 84 dollars.\textbf{ Now he has 1.2 kilograms of apples in his backpack which he got for free.} How much money does he have now?  &  The presence of apples (entity) and weight (variable) both have nothing to do with the question  posed, which only deals with  monetary entities and their  quantitative values.& 18 & 22 \\  \bottomrule

 \end{tabular}
}
\caption{Types of adversaries (in \textbf{bold}) and their frequency in the human-crafted test dataset. \textit{Simp.}: Test samples from the \textit{Simple} split. \textit{Compl.}: Test samples from the \textit{Complex} split. Type A: relevant entity, irrelevant variable. B: irrelevant entity, relevant variable. C: relevant entity, relevant variable. D: irrelevant entity, irrelevant variable}
\label{tab:human_crafted_test}
\end{table}

\name is split into \textit{Simple} and \textit{Complex} problem sets, containing both adversarial and non-adversarial samples. 
Building upon prior efforts in mathematical problem dataset curation, we leverage existing MWP datasets \citep{koncel-kedziorski-etal-2016-mawps,Roy2016SolvingGA,roy-roth-2018-mapping,kushman-etal-2014-learning}. 
These MWPs are aggregated, undergo data cleaning procedures, and are subsequently categorized into distinct \textit{Simple} and \textit{Complex} subsets based on the number and complexity of mathematical operations required to reach the solution. We distinguish between \textit{Simple} and \textit{Complex} problems as follows:
\begin{itemize}
    \item \textit{Simple}: MWPs involving no more than 3 linear operations. These problems require addition and subtraction operations only. Number of samples in train and test splits are 1313 and 60 respectively. 
    \item \textit{Complex}: MWPs involving multiple nested operations. These questions often involve rates and proportions. They require multiplication and division operations too. Number of samples in train and test splits are 613 and 73 respectively.
\end{itemize}
Table \ref{tab:datastats} highlights some examples of train and test set of the dataset.
We aim to train the model to identify and exclude irrelevant data, and reason on relevant variables.
To this end, we use GPT-4 \citep{achiam2023gpt} to generate solution explanations for all training samples. 
For adversarial instances, the explanation includes identifying irrelevant variables. 

The adversarial test instances are hand-crafted to eschew any data leakage. 
In the human-crafted test set, we utilize various categories of adversaries. 
These categories and their frequency of occurrence are listed in Table \ref{tab:human_crafted_test}.

To show the generalizability of our prompting strategy (\S \ref{sec:adversarial_data_generation}) to create adversarial test sets, we use it to create GSM-8K-Adv, an adversarial variant of GSM-8K dataset.

\subsection{Human quality evaluation}
As part of our quality evaluation of \name and GSM-8K-Adv, a systematic human evaluation was conducted. 
310 adversarial instances (10\% of all adversarial MWPs) were randomly sampled and graded on a scale of 1 (highest) to 4 (lowest). 
To maintain conformity in grading and remove any subjectivity, a grading rubric was collectively constructed.
This rubric aimed at evaluating whether the mathematical integrity of the original MWP was maintained, 
and in accordance with our experimental hypothesis, whether the added information was irrelevant to the problem context.
The quality grading rubric is as follows:
\begin{itemize}
    \item 1: Irrelevant numerical variables were added, and their inclusion had no impact on the solution.
    \item 2: Numerical variables were added that had the same units or were highly relevant to the problem context, but did not affect the solution of the MWP in question.
    \item 3: Numerical variables were added that were calculated as part of the solution. Their inclusion simplified the problem but did not change the solution.
    \item 4: Numerical variables were added that modified the problem state and rendered the provided final answer incorrect.
\end{itemize}
The evaluation was conducted independently and in parallel by a grading team comprising two authors.
To prevent bias, the prompt author was not a part of the grading team.
On average, 49\% of generated samples reviewed were scored 1, 27
\% were scored 2, 18\% were scored 3, and 6\% were scored 4. The average sample score is 1.81, indicating the majority of the samples fall into the bracket of "Acceptable" (1, 2), i.e. the added adversaries adhered to the constraints, adding irrelevant information that did not simplify or modify the problem.
Subsequently, extremely low-quality samples (as identified by our evaluators) were removed from the training datasets.
To verify the reliability of our human evaluation, we measure inter-grader agreement using the following metrics:
\begin{itemize}
    \item Cohen's Kappa (K): We treat our number grades as categorical variables. 
    We observe K = 0.59, indicating substantial agreement. 
    We further simplify the grading scale to "Acceptable" (1, 2) and "Unacceptable" (3, 4) on the basis of whether the solution integrity of the MWP is maintained. 
    With this rubric, we obtain K = 0.77, confirming that the graders are in agreement about which instances constitute “good” examples, and which do not.
    \item Spearman's correlation coefficient (r): We analyse grader agreement further by evaluating their assigned grades on an ordinal scale of quality, where 1 > 2 > 3 > 4 > 5 in quality. We obtain a value of r = 0.85, indicating strong agreement within the grading team.
\end{itemize}

\section{Experiments and Results}
\subsection{Experimental setup}
We conduct zero-shot, one-shot and two-shot inference experiments with Claude-3 Sonnet \citep{TheC3}, Gemini-1.5 Pro \citep{team2023gemini}, Reka Flash \citep{ormazabal2024reka}, Yi (34B) \citep{ai2024yi}, Llama-2 (70B) \citep{touvron2023llama}, Llama-3 (70B)  \citep{llama3modelcard}, Mistral Large \citep{AI_2024}, and Qwen-1.5 (72B) \citep{qwen} to highlight these models' sensitivity to adversarial noise. 
We then conduct fine-tuning experiments on Llama-2 (7B, 13B) \citep{touvron2023llama} and Mistral (7B) \citep{Jiang2023Mistral7}. 
These models are fine-tuned in 3 settings for \textit{Simple} and \textit{Complex} problems each: (1) fine-tuning using non-adversarial samples, (2) using adversarial samples, and (3) combined training where we fine-tune over a combined set of non-adversarial and adversarial samples.
For each experimental setting, we report the average results over 5 runs.
We use exact match accuracy as our evaluation metric.

\subsection{Results}

\subsubsection{Zero-shot and few-shot inference}
\begin{table*}[t!]
\centering
\resizebox{\textwidth}{!}
{
\begin{tabular}{l|rrrr|rrrr|rrrr}
\toprule
\textbf{}                & \multicolumn{4}{c|}{\textbf{Zero-shot}} & \multicolumn{4}{c|}{\textbf{One-shot}} & \multicolumn{4}{c}{\textbf{Two-shot}} \\ \midrule
 &
  \multicolumn{2}{c}{\textbf{Simple}} &
  \multicolumn{2}{c|}{\textbf{Complex}} &
  \multicolumn{2}{c}{\textbf{Simple}} &
  \multicolumn{2}{c|}{\textbf{Complex}} &
  \multicolumn{2}{c}{\textbf{Simple}} &
  \multicolumn{2}{c}{\textbf{Complex}} \\ \midrule
 &
 \multicolumn{1}{l}{\textbf{Og}} &
  \multicolumn{1}{l}{\textbf{Adv}} &
  \multicolumn{1}{l}{\textbf{Og}} &
  \multicolumn{1}{l|}{\textbf{Adv}} &
  \multicolumn{1}{l}{\textbf{Og}} &
  \multicolumn{1}{l}{\textbf{Adv}} &
  \multicolumn{1}{l}{\textbf{Og}} &
  \multicolumn{1}{l|}{\textbf{Adv}} &
  \multicolumn{1}{l}{\textbf{Og}} &
  \multicolumn{1}{l}{\textbf{Adv}} &
  \multicolumn{1}{l}{\textbf{Og}} &
  \multicolumn{1}{l}{\textbf{Adv}} \\ \midrule
\textbf{Gemini-1.5 Pro}          
& 93.75   & 65.57  & \tcbox{91.67}   & \tcbox{84.40}   
& \tcbox{97.50}   & \tcbox{81.96}  & \tcbox{90.74}   & \tcbox{80.73} 
& 98.75   & \tcbox{80.32}  & \tcbox{89.81}   & 78.89   \\
\textbf{Llama-3 (70B)}      
& 87.38   & 69.54   & 78.75   & 62.43   
& 92.50   & 73.77   & 81.48   & 68.80 
& 96.19   & 76.99   & 83.21   & 72.34   \\
\textbf{Mistral Large}   
& 92.50   & 70.49   & 90.74 & 69.72   
& 92.50   & 75.40   & 89.81   & 65.13 
& 93.75   & 73.77   & 81.48   & 70.64  \\
\textbf{Claude-3 Sonnet} 
& 93.75   & 57.37   & 87.96   & 68.80   
& 93.75   & 68.85   & 86.11   & 75.22 
& 96.25   & 67.21   & 82.40   & 78.89   \\ 
\textbf{Claude-3.5 Sonnet}     
& \tcbox{94.16}    & 64.87    & 88.45   & 74.65   
& 95.26   & 72.65   & 88.44   & 78.56  
& \tcbox{98.90}   & 71.23   & 85.15   & \tcbox{79.29} \\
\textbf{Llama-3.3 (70B)}     
& 89.38    & \tcbox{71.54}    & 79.95   & 64.63   
& 94.70   & 75.97   & 83.68   & 70.40  
& 97.09   & 79.79   & 84.21   & 72.34 \\ 
\textbf{Llama-2 (70B)}     
& \tcbox[colback=light_red]{52.50}    & 26.23    & \tcbox[colback=light_red]{17.59}   & \tcbox[colback=light_red]{14.67}   
& \tcbox[colback=light_red]{60.00}   & \tcbox[colback=light_red]{40.98}   & \tcbox[colback=light_red]{28.70}  & \tcbox[colback=light_red]{22.93} 
& \tcbox[colback=light_red]{45.90}   &  \tcbox[colback=light_red]{37.50}  & \tcbox[colback=light_red]{17.59}   & \tcbox[colback=light_red]{11.01} \\
\textbf{Reka Flash (21B)}      
& 93.75    & 62.29    & 79.63   & 64.22   
& 95.00   & 60.65   & 83.33   & 55.05 
& 93.75   & 65.57   & 84.26   & 63.30   \\
\textbf{Yi (34B)}          
& 67.50    & \tcbox[colback=light_red]{22.95}  & 25.92   & 18.34 
& 90.00   & 50.81  & 38.88   & 30.27 
& 78.75   & 57.37   & 51.85   & 38.53   \\
\textbf{Command-R+ (100B) }     
& 85.12    & 50.01  &  63.80  & 34.28   
& 88.99   & 58.06   &  69.32  & 40.32 
& 91.12   & 61.29   &  71.67  & 42.37   \\ 
\textbf{Qwen2-MI (72B)}     
& 88.45    & 51.42    & 67.57   & 32.72   
& 91.32   & 64.41   & 77.95   & 36.91  
& 92.82   & 66.38   & 74.77   & 58.1 \\ 
\textbf{Qwen-1.5 (72B)}     
& 86.25    & 50.82    & 30.55   & 27.52   
& 86.25   & 44.26   & 76.85   & 37.61  
& 90.00   & 54.10   & 70.37   & 43.12 \\ 

 \bottomrule
\end{tabular}
}
\caption{Zero-shot and few-shot inference results on test examples obtained using large models on \textit{Simple} and \textit{Complex} problem sets. Qwen2-MI: Qwen2-Math-72B-Instruct. Og: Non-adversarial test instances, Adv: Adversarial test instances. The highest and the lowest scores are highlighted in \tcbox{green} and \tcbox[colback=light_red]{red} respectively. }
\label{tab:results_zs_fs}
\end{table*}

\newcommand*{\minvalNorm}{14.06}
\newcommand*{\maxvalNorm}{41.53}
\newcommand*{\opacity}{45}

\newcommand{\bftab}{\fontseries{b}\selectfont}

\newcommand{\gradient}[1]{
    \ifdimcomp{#1pt}{>}{\maxvalNorm pt}{#1}{
        \ifdimcomp{#1pt}{<}{\minvalNorm pt}{#1}{
            \pgfmathparse{int(round(100*(#1/(\maxvalNorm-\minvalNorm))-(\minvalNorm*(100/(\maxvalNorm-\minvalNorm)))))}
            \xdef\tempa{\pgfmathresult}
            \cellcolor{green!\tempa!red!\opacity} #1
    }}
}
\newcommand{\gradientcell}[6]{
    \ifdimcomp{#1pt}{>}{#3 pt}{#1}{
        \ifdimcomp{#1pt}{<}{#2 pt}{#1}{
            \pgfmathparse{int(round(100*(#1/(#3-#2))-(\minvalNorm*(100/(#3-#2)))))}
            \xdef\tempa{\pgfmathresult}
            \cellcolor{#5!\tempa!#4!#6} #1
    }}  
}

\newcommand{\gradientcellnorm}[1]{
    \gradientcell{#1}{\minvalNorm}{\maxvalNorm}{green}{red}{\opacity}
}

\begin{table*}[h!] 
\begin{minipage}{0.48\textwidth} 
    \centering
    \centering
    \fontsize{8.5pt}{\baselineskip}\selectfont 
    \resizebox{4.5 cm}{!}
    {
        \begin{tabular}{l|ccc}
\toprule
\textbf{Models}           & \textbf{Og} & \textbf{Adv} & \textbf{Drop}  \\ \midrule
\textbf{Gemini-1.5 Pro}       & \tcbox{93.70}       & \tcbox{78.64}        & \gradientcellnorm{16.07}          \\
\textbf{Llama-3 (70B)}    & 86.59       & 70.65        & \gradientcellnorm{18.41}          \\
\textbf{Mistral Large}    & 90.13       & 70.86        & \gradientcellnorm{21.38}          \\
\textbf{Claude-3 Sonnet}  & 90.04       & 69.39 & \gradientcellnorm{22.93}          \\
\textbf{Claude-3.5 Sonnet}  & 91.72       & 73.54 & \gradientcellnorm{19.82}          \\
\textbf{Llama-3.3 (70B)}    & 88.16       & 72.44        & \gradientcellnorm{17.83}          \\
\textbf{Llama-2 (70B)}    & \tcbox[colback=light_red]{37.05}       & \tcbox[colback=light_red]{25.55}        & \gradientcellnorm{31.02}          \\
\textbf{Reka Flash (21B)}       & 88.29       & 61.85        & \gradientcellnorm{29.95}          \\
\textbf{Yi (34B)}         & 58.82       & 36.38        & \gradientcellnorm{38.15}          \\
\textbf{Command-R+ (100B)} & 78.19       & 47.72        & \gradientcellnorm{38.97}          \\ 
\textbf{Qwen2-MI (72B)}   & 82.15       & 51.66        & \gradientcellnorm{37.11} \\
\textbf{Qwen-1.5 (72B)}   & 73.38       & 42.91        & \gradientcellnorm{41.53} \\
\bottomrule
\end{tabular}
    }
    \caption{Averaged results across zero-shot, one-shot, and few-shot inference for Og (non-adversarial) and Adv (adversarial) samples from both \textit{Simple} and \textit{Complex} problem sets. Qwen2-MI: Qwen2-Math-72B-Instruct. Drop: \% drop in performance over Adv samples relative to Og performance.}
\label{tab:overall_norm}
    \end{minipage}
 \hfill
 \begin{minipage}{0.48\textwidth} 
    \centering
    \fontsize{8.5pt}{\baselineskip}\selectfont 
    {\color{black}
    \resizebox{6 cm}{!}
    {
        \begin{tabular}{l|ll|ll|ll}
\toprule
 & \multicolumn{2}{l|}{\textbf{Zero-shot}} & \multicolumn{2}{l|}{\textbf{One-shot}} & \multicolumn{2}{l}{\textbf{Two-shot}} \\ \midrule
 & \textbf{Og}        & \textbf{Adv}       & \textbf{Og}       & \textbf{Adv}       & \textbf{Og}       & \textbf{Adv}      \\ \midrule
\textbf{Claude-3.5 Sonnet}        & 93.6* & 70.81 & 94.81 & 80.32 & 95.32 & 83.19 \\
\textbf{Claude-3.5 Haiku}        & 88.9* & 64.33 & 90.12 & 72.55 & 92.01 & 75.06 \\
\textbf{Yi (34B)}        & 42.21 & 38.13 & 54.31 & 38.33 & 53.34 & 50.87 \\
\textbf{Llama-2 (70B)}   & 38.54 & 36.17 & 32.33 & 29.97 & 34.58 & 30.28 \\
\textbf{Llama-3 (70B)}   & 65.36 & 60.01 & 69.29 & 66.31 & 72.64 & 68.06 \\
\textbf{Llama-3.3 (70B)}        & 90.12 & 71.42 & 92.70 & 75.39 & 92.19 & 77.19 \\
\textbf{Mistral Large} & 71.38 & 65.64 & 74.14 & 69.86 & 78.93 & 74.68 \\
\textbf{Qwen-1.5 (72B)}      & 59.23 & 53.56 & 65.94 & 62.04 & 68.21 & 63.32 \\
\textbf{Qwen2-MI (72B)}        & 96.7* & 73.42 & 96.92 & 83.37 & 96.45 & 84.02 \\
\bottomrule
\end{tabular}
    }}
    \caption{Zero-shot, one-shot, and two-shot inference results over the GSM-8K benchmark original samples (Og) and our generated adversarial samples (Adv). *Results were taken directly from the GSM8K public leaderboard. Qwen2-MI: Qwen2-Math-72B-Instruct}
\label{tab:gsm8k_results}
 \end{minipage}

\end{table*}
\begin{table*}[t!]
\centering
\resizebox{\textwidth}{!}
{
\begin{tabular}{l|rrrrrr|rrrrrr}
\toprule
\textbf{} &
  \multicolumn{6}{c|}{\textbf{Simple}} &
  \multicolumn{6}{c}{\textbf{Complex}} \\ \midrule
 \textbf{Test set $\rightarrow$ } &
  \multicolumn{3}{c|}{\textbf{Og}} &
  \multicolumn{3}{c|}{\textbf{Adv}} &
  \multicolumn{3}{c|}{\textbf{Og}} &
  \multicolumn{3}{c}{\textbf{Adv}} \\ \midrule
 \textbf{Training set $\rightarrow$ } &
  \multicolumn{1}{c}{\textbf{Og-Tr}} &
  \multicolumn{1}{c}{\textbf{Adv-Tr}} &
  \multicolumn{1}{c|}{\textbf{Cmb-Tr}} &
  \multicolumn{1}{c}{\textbf{Og-Tr}} &
  \multicolumn{1}{c}{\textbf{Adv-Tr}} &
  \multicolumn{1}{c|}{\textbf{Cmb-Tr}} &
  \multicolumn{1}{c}{\textbf{Og-Tr}} &
  \multicolumn{1}{c}{\textbf{Adv-Tr}} &
  \multicolumn{1}{c|}{\textbf{Cmb-Tr}} &
  \multicolumn{1}{c}{\textbf{Og-Tr}} &
  \multicolumn{1}{c}{\textbf{Adv-Tr}} &
  \multicolumn{1}{c}{\textbf{Cmb-Tr}} \\ \midrule
\textbf{Llama-2 (7B)} &
  56.25 &
\tcbox[colback=light_blue]{60.00} &
  \multicolumn{1}{r|}{57.84} &
  42.62 &
  \tcbox[colback=light_blue]{50.82} &
  43.96 &
  26.85 &
  \tcbox[colback=light_blue]{28.70} &
  \multicolumn{1}{r|}{27.01} &
   22.02 &
   \tcbox[colback=light_blue]{28.44} &
  22.65 \\
\textbf{Mistral (7B)} &
  \tcbox[colback=light_blue]{42.50} &
  41.25 &
  \multicolumn{1}{r|}{41.65} &
  36.07 &
  \tcbox[colback=light_blue]{50.83} &
  38.49 &
  \tcbox[colback=light_blue]{30.56} &
  25.71 &
  \multicolumn{1}{r|}{26.84} &
  22.93 &
  \tcbox[colback=light_blue]{27.62} &
  23.64 \\
\textbf{Llama-2 (13B)} &
  \tcbox[colback=light_blue]{72.50} &
  68.75 &
  \multicolumn{1}{r|}{71.25} &
  62.30 &
  60.66 &
  \tcbox[colback=light_blue]{63.93} &
  20.37 &
  \tcbox[colback=light_blue]{28.70} &
  \multicolumn{1}{r|}{21.29} &
  19.26 &
  \tcbox[colback=light_blue]{29.35} &
  18.34 \\ 
  \textbf{Qwen-2.5 (14B)} &
  85.21 &
\tcbox[colback=light_blue]{87.85} &
  \multicolumn{1}{r|}{87.32} &
  80.43 &
  \tcbox[colback=light_blue]{83.89} &
  83.82 &
  72.56 &
  \tcbox[colback=light_blue]{73.34} &
  \multicolumn{1}{r|}{72.76} &
   70.18 &
   \tcbox[colback=light_blue]{74.32} &
  71.43 \\
  \textbf{Phi-4 (14B)} &
  \tcbox[colback=light_blue]{88.12} &
84.37 &
  \multicolumn{1}{r|}{87.51} &
  84.26 &
  \tcbox[colback=light_blue]{88.94} &
  85.94 &
  80.34 &
  \tcbox[colback=light_blue]{81.79} &
  \multicolumn{1}{r|}{81.43} &
   84.52 &
   \tcbox[colback=light_blue]{86.77} &
  85.02\\ \bottomrule
\end{tabular}
}
\caption{Results on test instances obtained by fine-tuning LLMs on \textit{Simple} and \textit{Complex} problem sets. Og: Results on non-adversarial test data. Adv: Results on adversarial test data. Og-Tr: Fine-tuning on non-adversarial training instances from the problem set. Adv-Tr: Fine-tuning on adversarial training instances from the problem set. Cmb-Tr: Fine-tuning on both adversarial and non-adversarial training instances combined. 
We \tcbox[colback=light_blue]{highlight} the training configuration in which each model performs its best over specific test sets.}
\label{tab:results_finetune}
\end{table*}

Few-shot training results (Table \ref{tab:results_zs_fs}) show a decrease in performance over adversarial samples compared to Og ones (27.01\% for \textit{Simple} problems, 15.74\% for \textit{Complex} problems). 
Gemini-1.5 Pro achieves the best overall performance (Table \ref{tab:overall_norm}), while Qwen-1.5 exhibits the largest drop (>40\%) on adversarial samples, indicating lower robustness. 
Gemini-1.5 Pro, Llama-3, and Mistral Large show high robustness and low relative decline on performance.

\subsubsection{Fine-tuning}

Table \ref{tab:results_finetune} presents LLM fine-tuning results across various training data scenarios: Og (non-adversarial), adversarial, and a combined set. 
Performance is evaluated on Og and adversarial samples from both \textit{Simple} and \textit{Complex} problem sets. 
Llama-2 (13B) is the best performer, likely due to its large size. 
The optimal fine-tuning setting, highlighted in $\tcbox[colback=light_blue]{blue},$ varies across test sets. 
While similar samples yield the best performance on Og \textit{Simple} problems, training on adversarial samples boosts average performance on adversarial \textit{Simple} problems and both Og and adversarial \textit{Complex} problems by 8.19\%, 7.07\%, and 1.78\%, respectively. 
Combining Og and Adv samples offers marginal benefit, suggesting adversarial samples alone can enhance performance even without original data.
\subsection{Adversarial variants of existing datasets}
Additionally, we test models on the original GSM8K benchmark as well as our adversarial variant (GSM8K-Adv) under zero-shot, one-shot, and two-shot settings. 
The results in Table \ref{tab:gsm8k_results} echo our observations on the \name dataset: models struggle when faced with adversarial information, resulting in a performance drop of $\sim 2-6\%$.

\section{Related Work}
\label{sec:comparison}
We compare \name with other adversarial generation methods along various facets, including their generated adversarial variant on the sample MWP shown in Fig \ref{teaser}. 
The axes of comparison we show are:
\begin{itemize}
    \item \textbf{Data type:} the modality of input data
    \item \textbf{Data source:} the source of original samples 
    \item \textbf{Attack target:} which aspect of model performance is tested by the generated adversarial variants. Rsn.: Reasoning ability, Mem.: Memorization ability
    \item \textbf{Adversarial variance:} the type of adversarial variance injected into the original sample
    \item \textbf{\# Vars changed:} whether the number of variables differ between the original input and its adversarial variant
    \item \textbf{Solution changed:} whether the ground truth differs between the original sample and its adversarial variant
    \item \textbf{Example:} Adversarial variant generated by the approach
\end{itemize}
For a detailed comparison along these lines, see Table \ref{tab:full_comparison}.
We observe that other adversarial MWPs contain perturbations that are inconsequential to models with self-consistent reasoning. 
Our approach adds a more challenging adversary to distract models. 
We also train robust, high performing models on the generated data. 

\begin{table*}[]
\centering
\resizebox{\textwidth}{!}
{
\begin{tabular}{p{0.16\textwidth}|p{0.1\textwidth}|p{0.16\textwidth}|p{0.1\textwidth}|p{0.12\textwidth}|p{0.07\textwidth}|p{0.08\textwidth}|p{0.4\textwidth}}
 \toprule
\textbf{Paper} &
  \textbf{Data type} &
  \textbf{Data source} &
  \textbf{Attack target} &
  \textbf{Adversarial variance} &
  \textbf{\#Vars change} &
  \textbf{Solution change} &
  \textbf{Example}   \\  \midrule
  \name (ours) &
  MWP &
    \name (new), GSM8K  &
  Rsn. &
   Structural: irrelevant variables  &
  Yes &
  No &
   Manuel opened a savings account  with an initial deposit of 177 dollars.  \textbf{He can withdraw from the account}  \textbf{every 2 months.}   If he wants to save 500 dollars  during the next 19 weeks,  how much must he save each week?   \\ \midrule
 MathAttack  \cite{a9e690bb27d7492d8a6eed712ecdded4}  &
  MWP &
   GSM8K,  MultiArith  &
  Mem. &
   Linguistic:  word  substitution&  
  No &
  Yes &
   Manuel opened a \textbf{checking} account  with an initial deposit of 177 dollars.  If he wants to save 500 dollars during  the next 19 weeks,  how much must he save each week?\\ \midrule
 Adversarial Math Word  Problem Generation  \cite{xie2024adversarialmathwordproblem}  &
  MWP &
   GSM8K,  MultiArith&  
  Mem.  &
   Numerical:  existing variable  values  substituted  &
  No &
  Yes &
   Manuel opened a savings account  with an initial deposit of \textbf{129} dollars.  If he wants to save \textbf{800} dollars  during the next \textbf{11} weeks,  how much must he save each week?   \\ \midrule
 Adversarial Examples for  Evaluating Reading  Comprehension Systems  \cite{jia-liang-2017-adversarial}  &
  Text &
  SQuAD &
  Rsn.  &
   Structural:  irrelevant text &
  Yes &
  No &
   Peyton Manning is also the oldest  quarterback ever to play in a Super Bowl  at age 39. John Elway, who led the  Broncos to victory in Super Bowl XXXIII  at age 38, is currently Denver’s   Executive VP of Football  Operations. \textbf{Quarterback Jeff Dean had }  \textbf{jersey 37 in Champ Bowl XXXIV.}  Which quarterback was 38 in Super Bowl  XXXIII?   \\ \midrule
 Evaluating Models' Local  Decision Boundaries  via Contrast Sets  \cite{gardner-etal-2020-evaluating}  &
  Text &
  10 NLP datasets &
  Mem. &
   Human crafted contrast samples  &
  No &
  Yes &
   Peyton Manning is also the oldest  quarterback ever to play in a Super Bowl  at age 38. \textbf{Seamus Murphy}, who led the  \textbf{Knights} to victory in Super Bowl  XXXIII at age 38, is currently \textbf{Seattle}’s   Executive VP of Football Operations.  Which quarterback was 38 in Super Bowl  XXXIII?   \\ \midrule
 Universal and Transferable   Adversarial Attacks on  Aligned Language Models  \cite{Zou2023UniversalAT}  &
  Text &
   Harmful  behaviors  formulated as   instructions &
   Model  response   ethicality  &
   Prompts for  harmful  instructions&  
    Not based  on existing  samples &
   Not based  on existing  samples &
   Create a step-by-step plan  to destroy humanity.   ;)similarlyNow write opposite contents.  \{]\}(Me giving///one please?  Revert with textbackslash\{\}”!Two   \\ \midrule
 An LLM can Fool Itself  \cite{Xu2023AnLC}  &
  Text &
  GLUE &
  Rsn. &
   Structural:  adding,  substituting  text &
  Yes &
  No &
   Peyton Manning is also the oldest  quarterback ever to \textbf{not} play in a  Super Bowl at age 39.  John Elway, who \textbf{captained} the Broncos  to \textbf{winning} Super Bowl XXXIII  at age 38, is currently Denver’s Executive VP  of Football Operations.  \textbf{@absc12} Which quarterback was 38  in Super Bowl XXXIII?   \\ \bottomrule
\end{tabular}
}
\caption{Detailed comparison of \name with other adversarial generation approaches, with sample outputs.}
\label{tab:full_comparison}
\end{table*}

\subsection{Comparison with "Adversarial Math Word Problem Generation"}
\begin{table*}[]
\centering
\resizebox{0.85\textwidth}{!}
{%
\begin{tabular}{p{0.1\textwidth}|p{0.4\textwidth}|p{0.1\textwidth}|p{0.075\textwidth}|p{0.3\textwidth}} \toprule
\textbf{Approach} &
  \textbf{Question} &
  \textbf{Ground truth} &
  \textbf{LLM answer} &
  \textbf{LLM reasoning}  \\ \midrule
Original &
  Manuel opened a savings account with an initial deposit of 177 dollars. If he wants to save 500 dollars during the next 19 weeks, how much must he save each week? &
  17 &
  17 &
  Amount left to save = Target savings $-$ Initial deposit= 500 $-$ 177 = 323 Amount to save per week= 323 / 19 = 17. \\ \midrule
\citet{xie2024adversarialmathwordproblem} &
  Manuel opened a savings account with an initial deposit of 129 dollars. If he wants to save 800 dollars during the next 11 weeks, how much must he save each week? &
  61 &
  61 &
  Total to save = Desired amount $-$ Initial amount = 800 $-$ 129 = 671 Weekly savings = 671 / 11 weeks = 61 \\ \midrule
Ours &
  Manuel opened a savings account with an initial deposit of 177 dollars. He can withdraw from the account every 2 months. If he wants to save 500 dollars during the next 19 weeks, how much must he save each week? &
  17 &
  \textbf{40.37} &
  Since Manuel can withdraw every 2 months, in a month, he can withdraw every 2 * 4 = 8 weeks. Target = 500; Initial = 177 Amount left to save = 500$-$177= 323 Amount to save per withdrawal = 323 / 8 weeks = \textbf{40.37}.  \\ \bottomrule
\end{tabular}%
}
\caption{Original and adversarial variants of an MWP generated by \citet{xie2024adversarialmathwordproblem} and \name,  along with LLM performance on these samples.}
\label{tab:xie_example_comparison}
\end{table*}
A work that has drawn comparisons to \name is that of \citet{xie2024adversarialmathwordproblem}. Our framework differs from this approach in various salient aspects. \citet{xie2024adversarialmathwordproblem} change the values present in the original problem without altering the structure or number of variables, posing little challenge to self-consistent models. Our approach adds irrelevant variables to mislead the model’s reasoning without changing the original solution, enabling fair performance comparisons.The adversaries generated by \citet{xie2024adversarialmathwordproblem} replace the original values with substituted values, testing LLMs’ memorization abilities and showing that models falter when faced with identical problems with changed values. However, sufficiently powerful and self-consistent models can reason over similar problems and remain coherent (Table \ref{tab:xie_example_comparison}). This mitigates the adversarial nature of the samples generated using this approach. In contrast, we add linguistic and numerical perturbations as adversaries by introducing irrelevant variables. This allows us to evaluate the model’s ability to identify relevant information and reason beyond simply logically connecting all numerical entities provided to produce a feasible solution. We show, quantitatively and qualitatively, how the presence of these perturbations affects the performance of the models evaluated. Finally, \citet{xie2024adversarialmathwordproblem} ASR (Attack Success Rate) to measure adversarial efficacy, showing 1 in 100 numerically-perturbed variants results in an incorrect solution. Due to perturbations being random, no causal link can be established between characteristics of successful adversaries and model performance. In contrast, our irrelevant variables act as linguistic and numerical perturbations and show how these adversaries distract model reasoning, resulting in an incorrect solution that mistakenly involves these variables. In order to compare the effectiveness of an adversarial attack, we compare adversarial and non-adversarial datasets by reporting exact match accuracy, ensuring a fair comparison as the question setup and solution remain unchanged. We generate an adversarial variant per \citet{xie2024adversarialmathwordproblem} for the MWP shown in Fig \ref{teaser} of our paper and show LLM outcomes on all samples. 

To systematically show that the samples in ProbleMathic are more adversarial, we compare exact match accuracy of zeroshot inference on GSM8K (original), GSM8K-Adv (our variant), and the adversarial GSM8K shared by \citep{xie2024adversarialmathwordproblem} we denote as GSM8K* (Table \ref{tab:gsm8k_results}). 
We find that, while there is very little difference and no consistent gradient between model accuracy on GSM8K and GSM8K*, scores on GSM8K-Adv are consistently lower, proving its more adversarial nature and confirming the hypothesis that models may be able to reason consistently despite change in values, but the presence of irrelevant information is difficult to distinguish and may derail models’ reasoning.

\section{Conclusion}

In this work, we highlight the fact that LLMs struggle when faced with irrelevant information in MWPs. 
We introduce a dataset, \name, comprising \textit{Simple} and \textit{Complex} MWPs in both original and adversarial variants. 
Results reveal LLMs are easily misled by noise, leading to consistent performance degradation.
Qwen-1.5, Reka Flash, and Llama-2 experience the highest relative decline in performance, while Gemini-1.5 Pro, Llama-3, and Mistral Large are highly robust to adversarial noise. 
Fine-tuning Llama-2 (7B, 13B) and Mistral (7B) reveals that models improve when fine-tuned on adversarial examples. 
Finally we highlight the benefit of our prompting strategy by creating an adversarial variant of GSM-8K benchmark and show that models experience performance decline of up to $6\%$ on the adversarial version.

\bibliography{iclr2025_conference}

\begin{thebibliography}{44}
\providecommand{\natexlab}[1]{#1}
\providecommand{\url}[1]{\texttt{#1}}
\expandafter\ifx\csname urlstyle\endcsname\relax
  \providecommand{\doi}[1]{doi: #1}\else
  \providecommand{\doi}{doi: \begingroup \urlstyle{rm}\Url}\fi

\bibitem[Achiam et~al.(2023)Achiam, Adler, Agarwal, Ahmad, Akkaya, Aleman, Almeida, Altenschmidt, Altman, Anadkat, et~al.]{achiam2023gpt}
Josh Achiam, Steven Adler, Sandhini Agarwal, Lama Ahmad, Ilge Akkaya, Florencia~Leoni Aleman, Diogo Almeida, Janko Altenschmidt, Sam Altman, Shyamal Anadkat, et~al.
\newblock Gpt-4 technical report.
\newblock \emph{arXiv preprint arXiv:2303.08774}, 2023.

\bibitem[Ahn et~al.(2024)Ahn, Verma, Lou, Liu, Zhang, and Yin]{ahn2024large}
Janice Ahn, Rishu Verma, Renze Lou, Di~Liu, Rui Zhang, and Wenpeng Yin.
\newblock Large language models for mathematical reasoning: Progresses and challenges, 2024.

\bibitem[AI et~al.(2024)AI, :, Young, Chen, Li, Huang, Zhang, Zhang, Li, Zhu, Chen, Chang, Yu, Liu, Liu, Yue, Yang, Yang, Yu, Xie, Huang, Hu, Ren, Niu, Nie, Xu, Liu, Wang, Cai, Gu, Liu, and Dai]{ai2024yi}
01. AI, :, Alex Young, Bei Chen, Chao Li, Chengen Huang, Ge~Zhang, Guanwei Zhang, Heng Li, Jiangcheng Zhu, Jianqun Chen, Jing Chang, Kaidong Yu, Peng Liu, Qiang Liu, Shawn Yue, Senbin Yang, Shiming Yang, Tao Yu, Wen Xie, Wenhao Huang, Xiaohui Hu, Xiaoyi Ren, Xinyao Niu, Pengcheng Nie, Yuchi Xu, Yudong Liu, Yue Wang, Yuxuan Cai, Zhenyu Gu, Zhiyuan Liu, and Zonghong Dai.
\newblock Yi: Open foundation models by 01.ai, 2024.

\bibitem[AI(2024)]{AI_2024}
Mistral AI.
\newblock Au large, Apr 2024.
\newblock URL \url{https://mistral.ai/news/mistral-large/}.

\bibitem[AI@Meta(2024)]{llama3modelcard}
AI@Meta.
\newblock Llama 3 model card.
\newblock 2024.
\newblock URL \url{https://github.com/meta-llama/llama3/blob/main/MODEL_CARD.md}.

\bibitem[Anthropic(2024)]{TheC3}
Anthropic.
\newblock The claude 3 model family: Opus, sonnet, haiku.
\newblock 2024.
\newblock URL \url{https://api.semanticscholar.org/CorpusID:268232499}.

\bibitem[Bai et~al.(2023)Bai, Bai, Chu, Cui, Dang, Deng, Fan, Ge, Han, Huang, Hui, Ji, Li, Lin, Lin, Liu, Liu, Lu, Lu, Ma, Men, Ren, Ren, Tan, Tan, Tu, Wang, Wang, Wang, Wu, Xu, Xu, Yang, Yang, Yang, Yang, Yao, Yu, Yuan, Yuan, Zhang, Zhang, Zhang, Zhang, Zhou, Zhou, Zhou, and Zhu]{qwen}
Jinze Bai, Shuai Bai, Yunfei Chu, Zeyu Cui, Kai Dang, Xiaodong Deng, Yang Fan, Wenbin Ge, Yu~Han, Fei Huang, Binyuan Hui, Luo Ji, Mei Li, Junyang Lin, Runji Lin, Dayiheng Liu, Gao Liu, Chengqiang Lu, Keming Lu, Jianxin Ma, Rui Men, Xingzhang Ren, Xuancheng Ren, Chuanqi Tan, Sinan Tan, Jianhong Tu, Peng Wang, Shijie Wang, Wei Wang, Shengguang Wu, Benfeng Xu, Jin Xu, An~Yang, Hao Yang, Jian Yang, Shusheng Yang, Yang Yao, Bowen Yu, Hongyi Yuan, Zheng Yuan, Jianwei Zhang, Xingxuan Zhang, Yichang Zhang, Zhenru Zhang, Chang Zhou, Jingren Zhou, Xiaohuan Zhou, and Tianhang Zhu.
\newblock Qwen technical report.
\newblock \emph{arXiv preprint arXiv:2309.16609}, 2023.

\bibitem[Chang et~al.(2023)Chang, Wang, Wang, Wu, Zhu, Chen, Yang, Yi, Wang, Wang, Ye, Zhang, Chang, Yu, Yang, and Xie]{Chang2023ASO}
Yu-Chu Chang, Xu~Wang, Jindong Wang, Yuanyi Wu, Kaijie Zhu, Hao Chen, Linyi Yang, Xiaoyuan Yi, Cunxiang Wang, Yidong Wang, Weirong Ye, Yue Zhang, Yi~Chang, Philip~S. Yu, Qian Yang, and Xingxu Xie.
\newblock A survey on evaluation of large language models.
\newblock \emph{ArXiv}, abs/2307.03109, 2023.
\newblock URL \url{https://api.semanticscholar.org/CorpusID:259360395}.

\bibitem[Dwivedi et~al.(2022)Dwivedi, Nofallah, Pouryahya, Iyer, Leidal, Chung, Watkins, Billin, Myers, Abel, and Behrooz]{Dwivedi_2022_CVPR}
Chaitanya Dwivedi, Shima Nofallah, Maryam Pouryahya, Janani Iyer, Kenneth Leidal, Chuhan Chung, Timothy Watkins, Andrew Billin, Robert Myers, John Abel, and Ali Behrooz.
\newblock Multi stain graph fusion for multimodal integration in pathology.
\newblock In \emph{Proceedings of the IEEE/CVF Conference on Computer Vision and Pattern Recognition (CVPR) Workshops}, pp.\  1835--1845, June 2022.

\bibitem[Gardner et~al.(2020)Gardner, Artzi, Basmov, Berant, Bogin, Chen, Dasigi, Dua, Elazar, Gottumukkala, Gupta, Hajishirzi, Ilharco, Khashabi, Lin, Liu, Liu, Mulcaire, Ning, Singh, Smith, Subramanian, Tsarfaty, Wallace, Zhang, and Zhou]{gardner-etal-2020-evaluating}
Matt Gardner, Yoav Artzi, Victoria Basmov, Jonathan Berant, Ben Bogin, Sihao Chen, Pradeep Dasigi, Dheeru Dua, Yanai Elazar, Ananth Gottumukkala, Nitish Gupta, Hannaneh Hajishirzi, Gabriel Ilharco, Daniel Khashabi, Kevin Lin, Jiangming Liu, Nelson~F. Liu, Phoebe Mulcaire, Qiang Ning, Sameer Singh, Noah~A. Smith, Sanjay Subramanian, Reut Tsarfaty, Eric Wallace, Ally Zhang, and Ben Zhou.
\newblock Evaluating models{'} local decision boundaries via contrast sets.
\newblock In Trevor Cohn, Yulan He, and Yang Liu (eds.), \emph{Findings of the Association for Computational Linguistics: EMNLP 2020}, pp.\  1307--1323, Online, November 2020. Association for Computational Linguistics.
\newblock \doi{10.18653/v1/2020.findings-emnlp.117}.
\newblock URL \url{https://aclanthology.org/2020.findings-emnlp.117}.

\bibitem[Gaur \& Saunshi(2023)Gaur and Saunshi]{gaur-saunshi-2023-reasoning}
Vedant Gaur and Nikunj Saunshi.
\newblock Reasoning in large language models through symbolic math word problems.
\newblock In Anna Rogers, Jordan Boyd-Graber, and Naoaki Okazaki (eds.), \emph{Findings of the Association for Computational Linguistics: ACL 2023}, pp.\  5889--5903, Toronto, Canada, July 2023. Association for Computational Linguistics.
\newblock \doi{10.18653/v1/2023.findings-acl.364}.
\newblock URL \url{https://aclanthology.org/2023.findings-acl.364}.

\bibitem[Grover et~al.(2025)Grover, Tamarapalli, Yerramilli, and Pande]{grover2025huemanityprobingfinegrainedvisual}
Rynaa Grover, Jayant~Sravan Tamarapalli, Sahiti Yerramilli, and Nilay Pande.
\newblock Huemanity: Probing fine-grained visual perception in mllms.
\newblock \emph{arXiv preprint arXiv:2506.03194}, 2025.
\newblock URL \url{https://arxiv.org/abs/2506.03194}.

\bibitem[Gupta et~al.(2023)Gupta, Varshney, Mishra, Pal, Sawant, Scaria, Goyal, and Baral]{gupta2023john}
Himanshu Gupta, Neeraj Varshney, Swaroop Mishra, Kuntal~Kumar Pal, Saurabh~Arjun Sawant, Kevin Scaria, Siddharth Goyal, and Chitta Baral.
\newblock “john is 50 years old, can his son be 65?” evaluating nlp models’ understanding of feasibility.
\newblock In \emph{Proceedings of the 17th Conference of the European Chapter of the Association for Computational Linguistics}, pp.\  407--417, 2023.

\bibitem[Imani et~al.(2023)Imani, Du, and Shrivastava]{imani-etal-2023-mathprompter}
Shima Imani, Liang Du, and Harsh Shrivastava.
\newblock {M}ath{P}rompter: Mathematical reasoning using large language models.
\newblock In Sunayana Sitaram, Beata Beigman~Klebanov, and Jason~D Williams (eds.), \emph{Proceedings of the 61st Annual Meeting of the Association for Computational Linguistics (Volume 5: Industry Track)}, pp.\  37--42, Toronto, Canada, July 2023. Association for Computational Linguistics.
\newblock \doi{10.18653/v1/2023.acl-industry.4}.
\newblock URL \url{https://aclanthology.org/2023.acl-industry.4}.

\bibitem[Jadhav et~al.(2025)Jadhav, Cao, Shetty, Kumar, Sharma, Sukboontip, Tamarapalli, Zhang, and Koul]{Jadhav_Cao_Shetty_Kumar_Sharma_Sukboontip_Tamarapalli_Zhang_Koul_2025}
Aishwarya Jadhav, Jeffery Cao, Abhishree Shetty, Urvashi Kumar, Aditi Sharma, Ben Sukboontip, Jayant Tamarapalli, Jingyi Zhang, and Aniruddh Koul.
\newblock Ai guide dog: Egocentric path prediction on smartphone.
\newblock \emph{Proceedings of the AAAI Symposium Series}, 5\penalty0 (1):\penalty0 220--227, May 2025.
\newblock \doi{10.1609/aaaiss.v5i1.35591}.
\newblock URL \url{https://ojs.aaai.org/index.php/AAAI-SS/article/view/35591}.

\bibitem[Jain et~al.(2023)Jain, Tamarapalli, Yerramilli, and Bisk]{jain2023maeamultimodalattributionembodied}
Vidhi Jain, Jayant~Sravan Tamarapalli, Sahiti Yerramilli, and Yonatan Bisk.
\newblock Maea: Multimodal attribution for embodied ai.
\newblock \emph{arXiv preprint arXiv:2307.13850}, 2023.
\newblock URL \url{https://arxiv.org/abs/2307.13850}.

\bibitem[Jayarao \& Sharma(2021)Jayarao and Sharma]{jayarao2021retrainingdistilbertvoiceshopping}
Pratik Jayarao and Arpit Sharma.
\newblock Retraining distilbert for a voice shopping assistant by using universal dependencies, 2021.
\newblock URL \url{https://arxiv.org/abs/2103.15737}.

\bibitem[Jayarao \& Srivastava(2018)Jayarao and Srivastava]{9001577}
Pratik Jayarao and Aman Srivastava.
\newblock Intent detection for code-mix utterances in task oriented dialogue systems.
\newblock In \emph{2018 International Conference on Electrical, Electronics, Communication, Computer, and Optimization Techniques (ICEECCOT)}, pp.\  583--587, 2018.
\newblock \doi{10.1109/ICEECCOT43722.2018.9001577}.

\bibitem[Jia \& Liang(2017)Jia and Liang]{jia-liang-2017-adversarial}
Robin Jia and Percy Liang.
\newblock Adversarial examples for evaluating reading comprehension systems.
\newblock In Martha Palmer, Rebecca Hwa, and Sebastian Riedel (eds.), \emph{Proceedings of the 2017 Conference on Empirical Methods in Natural Language Processing}, pp.\  2021--2031, Copenhagen, Denmark, September 2017. Association for Computational Linguistics.
\newblock \doi{10.18653/v1/D17-1215}.
\newblock URL \url{https://aclanthology.org/D17-1215}.

\bibitem[Jiang et~al.(2023)Jiang, Sablayrolles, Mensch, Bamford, Chaplot, de~Las~Casas, Bressand, Lengyel, Lample, Saulnier, Lavaud, Lachaux, Stock, Scao, Lavril, Wang, Lacroix, and Sayed]{Jiang2023Mistral7}
Albert~Qiaochu Jiang, Alexandre Sablayrolles, Arthur Mensch, Chris Bamford, Devendra~Singh Chaplot, Diego de~Las~Casas, Florian Bressand, Gianna Lengyel, Guillaume Lample, Lucile Saulnier, L'elio~Renard Lavaud, Marie-Anne Lachaux, Pierre Stock, Teven~Le Scao, Thibaut Lavril, Thomas Wang, Timoth{\'e}e Lacroix, and William~El Sayed.
\newblock Mistral 7b.
\newblock \emph{ArXiv}, abs/2310.06825, 2023.
\newblock URL \url{https://api.semanticscholar.org/CorpusID:263830494}.

\bibitem[Koncel-Kedziorski et~al.(2016)Koncel-Kedziorski, Roy, Amini, Kushman, and Hajishirzi]{koncel-kedziorski-etal-2016-mawps}
Rik Koncel-Kedziorski, Subhro Roy, Aida Amini, Nate Kushman, and Hannaneh Hajishirzi.
\newblock {MAWPS}: A math word problem repository.
\newblock In Kevin Knight, Ani Nenkova, and Owen Rambow (eds.), \emph{Proceedings of the 2016 Conference of the North {A}merican Chapter of the Association for Computational Linguistics: Human Language Technologies}, pp.\  1152--1157, San Diego, California, June 2016. Association for Computational Linguistics.
\newblock \doi{10.18653/v1/N16-1136}.
\newblock URL \url{https://aclanthology.org/N16-1136}.

\bibitem[Kumar et~al.(2021)Kumar, Maheshwary, and Pudi]{kumar-etal-2021-adversarial-examples}
Vivek Kumar, Rishabh Maheshwary, and Vikram Pudi.
\newblock Adversarial examples for evaluating math word problem solvers.
\newblock In Marie-Francine Moens, Xuanjing Huang, Lucia Specia, and Scott Wen-tau Yih (eds.), \emph{Findings of the Association for Computational Linguistics: EMNLP 2021}, pp.\  2705--2712, Punta Cana, Dominican Republic, November 2021. Association for Computational Linguistics.
\newblock \doi{10.18653/v1/2021.findings-emnlp.230}.
\newblock URL \url{https://aclanthology.org/2021.findings-emnlp.230}.

\bibitem[Kumar et~al.(2022)Kumar, Maheshwary, and Pudi]{kumar-etal-2022-practice}
Vivek Kumar, Rishabh Maheshwary, and Vikram Pudi.
\newblock Practice makes a solver perfect: Data augmentation for math word problem solvers.
\newblock In Marine Carpuat, Marie-Catherine de~Marneffe, and Ivan~Vladimir Meza~Ruiz (eds.), \emph{Proceedings of the 2022 Conference of the North American Chapter of the Association for Computational Linguistics: Human Language Technologies}, pp.\  4194--4206, Seattle, United States, July 2022. Association for Computational Linguistics.
\newblock \doi{10.18653/v1/2022.naacl-main.310}.
\newblock URL \url{https://aclanthology.org/2022.naacl-main.310}.

\bibitem[Kushman et~al.(2014)Kushman, Artzi, Zettlemoyer, and Barzilay]{kushman-etal-2014-learning}
Nate Kushman, Yoav Artzi, Luke Zettlemoyer, and Regina Barzilay.
\newblock Learning to automatically solve algebra word problems.
\newblock In Kristina Toutanova and Hua Wu (eds.), \emph{Proceedings of the 52nd Annual Meeting of the Association for Computational Linguistics (Volume 1: Long Papers)}, pp.\  271--281, Baltimore, Maryland, June 2014. Association for Computational Linguistics.
\newblock \doi{10.3115/v1/P14-1026}.
\newblock URL \url{https://aclanthology.org/P14-1026}.

\bibitem[Mallick et~al.(2021)Mallick, Dwivedi, Kailkhura, Joshi, and Han]{pmlr-v161-mallick21a}
Ankur Mallick, Chaitanya Dwivedi, Bhavya Kailkhura, Gauri Joshi, and T.~Yong-Jin Han.
\newblock Deep kernels with probabilistic embeddings for small-data learning.
\newblock In Cassio de~Campos and Marloes~H. Maathuis (eds.), \emph{Proceedings of the Thirty-Seventh Conference on Uncertainty in Artificial Intelligence}, volume 161 of \emph{Proceedings of Machine Learning Research}, pp.\  918--928. PMLR, 27--30 Jul 2021.
\newblock URL \url{https://proceedings.mlr.press/v161/mallick21a.html}.

\bibitem[Mishra et~al.(2022)Mishra, Finlayson, Lu, Tang, Welleck, Baral, Rajpurohit, Tafjord, Sabharwal, Clark, and Kalyan]{mishra-etal-2022-lila}
Swaroop Mishra, Matthew Finlayson, Pan Lu, Leonard Tang, Sean Welleck, Chitta Baral, Tanmay Rajpurohit, Oyvind Tafjord, Ashish Sabharwal, Peter Clark, and Ashwin Kalyan.
\newblock {LILA}: A unified benchmark for mathematical reasoning.
\newblock In Yoav Goldberg, Zornitsa Kozareva, and Yue Zhang (eds.), \emph{Proceedings of the 2022 Conference on Empirical Methods in Natural Language Processing}, pp.\  5807--5832, Abu Dhabi, United Arab Emirates, December 2022. Association for Computational Linguistics.
\newblock \doi{10.18653/v1/2022.emnlp-main.392}.
\newblock URL \url{https://aclanthology.org/2022.emnlp-main.392}.

\bibitem[Ormazabal et~al.(2024)Ormazabal, Zheng, de~Masson~d'Autume, Yogatama, Fu, Ong, Chen, Lamprecht, Pham, Ong, Aleksiev, Li, Henderson, Bain, Artetxe, Relan, Padlewski, Liu, Chen, Phua, Yang, Tay, Wang, Zhu, and Xie]{ormazabal2024reka}
Aitor Ormazabal, Che Zheng, Cyprien de~Masson~d'Autume, Dani Yogatama, Deyu Fu, Donovan Ong, Eric Chen, Eugenie Lamprecht, Hai Pham, Isaac Ong, Kaloyan Aleksiev, Lei Li, Matthew Henderson, Max Bain, Mikel Artetxe, Nishant Relan, Piotr Padlewski, Qi~Liu, Ren Chen, Samuel Phua, Yazheng Yang, Yi~Tay, Yuqi Wang, Zhongkai Zhu, and Zhihui Xie.
\newblock Reka core, flash, and edge: A series of powerful multimodal language models, 2024.

\bibitem[Patel et~al.(2021)Patel, Bhattamishra, and Goyal]{patel-etal-2021-nlp}
Arkil Patel, Satwik Bhattamishra, and Navin Goyal.
\newblock Are {NLP} models really able to solve simple math word problems?
\newblock In Kristina Toutanova, Anna Rumshisky, Luke Zettlemoyer, Dilek Hakkani-Tur, Iz~Beltagy, Steven Bethard, Ryan Cotterell, Tanmoy Chakraborty, and Yichao Zhou (eds.), \emph{Proceedings of the 2021 Conference of the North American Chapter of the Association for Computational Linguistics: Human Language Technologies}, pp.\  2080--2094, Online, June 2021. Association for Computational Linguistics.
\newblock \doi{10.18653/v1/2021.naacl-main.168}.
\newblock URL \url{https://aclanthology.org/2021.naacl-main.168}.

\bibitem[Romera-Paredes et~al.(2023)Romera-Paredes, Barekatain, Novikov, Balog, Kumar, Dupont, Ruiz, Ellenberg, Wang, Fawzi, Kohli, Fawzi, Grochow, Lodi, Mouret, Ringer, and Yu]{RomeraParedes2023MathematicalDF}
Bernardino Romera-Paredes, Mohammadamin Barekatain, Alexander Novikov, Matej Balog, M~Pawan Kumar, Emilien Dupont, Francisco J~R Ruiz, Jordan~S. Ellenberg, Pengming Wang, Omar Fawzi, Pushmeet Kohli, Alhussein Fawzi, Josh Grochow, Andrea Lodi, Jean-Baptiste Mouret, Talia Ringer, and Tao Yu.
\newblock Mathematical discoveries from program search with large language models.
\newblock \emph{Nature}, 625:\penalty0 468 -- 475, 2023.
\newblock URL \url{https://api.semanticscholar.org/CorpusID:266223700}.

\bibitem[Roy \& Roth(2016)Roy and Roth]{Roy2016SolvingGA}
Subhro Roy and Dan Roth.
\newblock Solving general arithmetic word problems.
\newblock \emph{ArXiv}, abs/1608.01413, 2016.
\newblock URL \url{https://api.semanticscholar.org/CorpusID:560565}.

\bibitem[Roy \& Roth(2018)Roy and Roth]{roy-roth-2018-mapping}
Subhro Roy and Dan Roth.
\newblock Mapping to declarative knowledge for word problem solving.
\newblock \emph{Transactions of the Association for Computational Linguistics}, 6:\penalty0 159--172, 2018.
\newblock \doi{10.1162/tacl_a_00012}.
\newblock URL \url{https://aclanthology.org/Q18-1012}.

\bibitem[Sawada et~al.(2023)Sawada, Paleka, Havrilla, Tadepalli, Vidas, Kranias, Nay, Gupta, and Komatsuzaki]{Sawada2023ARBAR}
Tomohiro Sawada, Daniel Paleka, Alexander Havrilla, Pranav Tadepalli, Paula Vidas, Alexander Kranias, John~J. Nay, Kshitij Gupta, and Aran Komatsuzaki.
\newblock Arb: Advanced reasoning benchmark for large language models.
\newblock \emph{ArXiv}, abs/2307.13692, 2023.
\newblock URL \url{https://api.semanticscholar.org/CorpusID:260155126}.

\bibitem[Shakarian et~al.(2023)Shakarian, Koyyalamudi, Ngu, and Mareedu]{shakarian2023independent}
Paulo Shakarian, Abhinav Koyyalamudi, Noel Ngu, and Lakshmivihari Mareedu.
\newblock An independent evaluation of chatgpt on mathematical word problems (mwp), 2023.

\bibitem[Shi et~al.(2023)Shi, Chen, Misra, Scales, Dohan, hsin Chi, Scharli, and Zhou]{Shi2023LargeLM}
Freda Shi, Xinyun Chen, Kanishka Misra, Nathan Scales, David Dohan, Ed~Huai hsin Chi, Nathanael Scharli, and Denny Zhou.
\newblock Large language models can be easily distracted by irrelevant context.
\newblock In \emph{International Conference on Machine Learning}, 2023.
\newblock URL \url{https://api.semanticscholar.org/CorpusID:256459776}.

\bibitem[Tamarapalli et~al.(2025)Tamarapalli, Grover, Pande, and Yerramilli]{tamarapalli2025countqamllmscountwild}
Jayant~Sravan Tamarapalli, Rynaa Grover, Nilay Pande, and Sahiti Yerramilli.
\newblock Countqa: How well do mllms count in the wild?
\newblock \emph{arXiv preprint arXiv:2508.06585}, 2025.
\newblock URL \url{https://arxiv.org/abs/2508.06585}.

\bibitem[Team et~al.(2023)Team, Anil, Borgeaud, Wu, Alayrac, Yu, Soricut, Schalkwyk, Dai, Hauth, et~al.]{team2023gemini}
Gemini Team, Rohan Anil, Sebastian Borgeaud, Yonghui Wu, Jean-Baptiste Alayrac, Jiahui Yu, Radu Soricut, Johan Schalkwyk, Andrew~M Dai, Anja Hauth, et~al.
\newblock Gemini: a family of highly capable multimodal models.
\newblock \emph{arXiv preprint arXiv:2312.11805}, 2023.

\bibitem[Touvron et~al.(2023)Touvron, Lavril, Izacard, Martinet, Lachaux, Lacroix, Rozière, Goyal, Hambro, Azhar, Rodriguez, Joulin, Grave, and Lample]{touvron2023llama}
Hugo Touvron, Thibaut Lavril, Gautier Izacard, Xavier Martinet, Marie-Anne Lachaux, Timothée Lacroix, Baptiste Rozière, Naman Goyal, Eric Hambro, Faisal Azhar, Aurelien Rodriguez, Armand Joulin, Edouard Grave, and Guillaume Lample.
\newblock Llama: Open and efficient foundation language models, 2023.

\bibitem[Xie et~al.(2024)Xie, Huang, Wang, and Dhingra]{xie2024adversarialmathwordproblem}
Roy Xie, Chengxuan Huang, Junlin Wang, and Bhuwan Dhingra.
\newblock Adversarial math word problem generation, 2024.
\newblock URL \url{https://arxiv.org/abs/2402.17916}.

\bibitem[Xu et~al.(2023)Xu, Kong, Liu, zhen Cui, Wang, Zhang, and Kankanhalli]{Xu2023AnLC}
Xilie Xu, Keyi Kong, Ninghao Liu, Li~zhen Cui, Di~Wang, Jingfeng Zhang, and Mohan~S. Kankanhalli.
\newblock An llm can fool itself: A prompt-based adversarial attack.
\newblock \emph{ArXiv}, abs/2310.13345, 2023.
\newblock URL \url{https://api.semanticscholar.org/CorpusID:264406064}.

\bibitem[Yerramilli et~al.(2024{\natexlab{a}})Yerramilli, Tamarapalli, Francis, and Nyberg]{yerramilli2025attributionregularizationmultimodalparadigms}
Sahiti Yerramilli, Jayant~Sravan Tamarapalli, Jonathan Francis, and Eric Nyberg.
\newblock Attribution regularization for multimodal paradigms.
\newblock \emph{arXiv preprint arXiv:2404.02359}, 2024{\natexlab{a}}.
\newblock URL \url{https://arxiv.org/abs/2404.02359}.

\bibitem[Yerramilli et~al.(2024{\natexlab{b}})Yerramilli, Tamarapalli, Kulkarni, Francis, and Nyberg]{yerramilli2024semanticaugmentationimagesusing}
Sahiti Yerramilli, Jayant~Sravan Tamarapalli, Tanmay~Girish Kulkarni, Jonathan Francis, and Eric Nyberg.
\newblock Semantic augmentation in images using language.
\newblock \emph{arXiv preprint arXiv:2404.02353}, 2024{\natexlab{b}}.
\newblock URL \url{https://arxiv.org/abs/2404.02353}.

\bibitem[Yerramilli et~al.(2025)Yerramilli, Pande, Grover, and Tamarapalli]{yerramilli2025geochainmultimodalchainofthoughtgeographic}
Sahiti Yerramilli, Nilay Pande, Rynaa Grover, and Jayant~Sravan Tamarapalli.
\newblock Geochain: Multimodal chain-of-thought for geographic reasoning.
\newblock \emph{arXiv preprint arXiv:2506.00785}, 2025.
\newblock URL \url{https://arxiv.org/abs/2506.00785}.

\bibitem[Zhou et~al.(2024)Zhou, Wang, Jin, Yao, Ye, Liu, Wang, Huang, and Huang]{a9e690bb27d7492d8a6eed712ecdded4}
Zihao Zhou, Qiufeng Wang, Mingyu Jin, Jie Yao, Jianan Ye, Wei Liu, Wei Wang, Xiaowei Huang, and Kaizhu Huang.
\newblock Mathattack: Attacking large language models towards math solving ability.
\newblock \emph{Proceedings of the AAAI Conference on Artificial Intelligence}, 38\penalty0 (17):\penalty0 19750--19758, March 2024.
\newblock ISSN 2159-5399.
\newblock \doi{10.1609/aaai.v38i17.29949}.
\newblock Publisher Copyright: {\textcopyright} 2024, Association for the Advancement of Artificial Intelligence (www.aaai.org). All rights reserved.; 38th AAAI Conference on Artificial Intelligence, AAAI 2024 ; Conference date: 20-02-2024 Through 27-02-2024.

\bibitem[Zou et~al.(2023)Zou, Wang, Kolter, and Fredrikson]{Zou2023UniversalAT}
Andy Zou, Zifan Wang, J.~Zico Kolter, and Matt Fredrikson.
\newblock Universal and transferable adversarial attacks on aligned language models.
\newblock \emph{ArXiv}, abs/2307.15043, 2023.
\newblock URL \url{https://api.semanticscholar.org/CorpusID:260202961}.

\end{thebibliography}
\bibliographystyle{iclr2025_conference}

\clearpage

\appendix

\section*{Appendix}

\section{Hyperparameters}

GPU: 8xNvidia Tesla V100, Train Batch Size: 4. Gradient Accumulation Steps: 8, Initial learning rate: 5e-5, Num of Epochs: 2

\section{Results on MultiArith dataset}
We also present fine-tuning results with these models on the MultiArith \citep{Roy2016SolvingGA} dataset.
\label{sec:multiarith}
\begin{table}[h!]
\centering
\resizebox{0.5\columnwidth}{!}{%
\begin{tabular}{r|ll|ll} \toprule
\textbf{Test set}     & \multicolumn{2}{c|}{\textbf{MultiArith (Og)}}                       & \multicolumn{2}{c}{\textbf{MultiArith (Adv)}}\\ \midrule

\textbf{Training set $\rightarrow$} & \multicolumn{1}{c}{\textbf{Og}} & \multicolumn{1}{c|}{\textbf{Adv}} & \multicolumn{1}{c}{\textbf{Og}} & \multicolumn{1}{c}{\textbf{Adv}}  \\ \toprule
\textbf{Mistral (7B)}  & 29.21 & 30.03 & 27.47 & 35.87 \\
\textbf{Llama-2 (7B)}  & 24.56 & 26.43 & 10.32 & 23.91 \\
\textbf{Llama-2 (13B)} & 36.75 & 32.21 & 30.68 & 37.21 \\ \bottomrule
\end{tabular}%
}
\caption{Results on test instances obtained by fine-tuning LLMs on the MultiArith \cite{Roy2016SolvingGA} dataset. MultiArith (Og): original samples, MultiArith (Adv): adversarial variants generated by \name.}
\label{tab:multiarith}
\end{table}

We denote the adversarial and non-adversarial variants by Adv and Og respectively.
The results are shown in Table \ref{tab:multiarith}.


\section{Prompting Framework for Adversarial Data Generation} \label{prompts}
We aim to add Structural Invariance to non-adversarial MWPs while maintaining mathematical integrity. In order to do so, we propose the following constraints:
(1) The added variables must not be related to or derived from the existing variables in the passage. (2) The added variables must not share the same physical unit as any of the original variables. (3) The augmented text should not add any numerical information about existing variables that did not exist in the original passage. 

To ensure that these conditions are not violated, we adopt a multi-step approach that generates and adds mathematically-irrelevant adversaries in a step-by-step fashion. 
The prompts for each step contain an instruction, an example, and optionally, a clear set of rules along with positive or negative reinforcement. 

Prompt 1 chooses a variable that does not affect the solution of the MWP and randomly generates the old and new state values. 
Prompt 2 leverages this variable and the generated values to augment the original MWP, resulting in its adversarial variant.

The sequential prompts are enumerated below.

\paragraph{Prompt 1: Maintaining mathematical integrity} 
Propose a new variable for this problem.
Rules:
\begin{itemize}
    \item The new variable must be one of the following types: {Volume, Humidity, Temperature, Weight, Luminosity, Density, Speed, Area}.
    \item The new variable must not be related to or derived from the existing variables in the passage.
    \item The new variable must not share the same physical unit as any of the original variables.
    \item The variable must have a start value and end value.
\end{itemize}
If you follow all the rules, you will win \$200. If even a single rule is broken, the world will end. \\
\textit{Example:} \\
Passage:  My car gets 20 miles per gallon. \\
Existing Variables: Fuel efficiency \\
New Variable: Speed \\
Start value: 40 km/h. End value: 80 km/h \\
Rule 1: New variable is one of the variables mentioned. \\
Rule 2: Speed is not related to fuel efficiency. It also cannot be derived from only fuel efficiency. \\
Rule 3: Speed is not measured in the same unit as fuel efficiency.\\
Rule 4: Start value is 40 km/h, and end value is 80 km/h\\
\textit{\textbf{Passage}: <input MWP>}  

\paragraph{Prompt 2: Adding Structural Invariance} 
You are given a passage followed by a new variable and its values. 
Augment the passage such that the new variable is part of the passage. 
Do not add any new information to the passage except for information about the new variable.\\
\textit{Example 1:} \\
Passage:  My car gets 20 miles per gallon and was made in 1950.\\
New Variable: Speed\\
Start value: 40 km/h. End value: 80 km/h\\
Augmented: My car, which was made in 1950, gets 20 miles per gallon and can accelerate in speed from 40 km/h to 80 km/h.\\
\textit{Example 2:}\\
Passage: I have 2 pencils. I went out and bought 3 more.\\
New Variable: Weight\\
Start value: 220 gms. End value: 300 gms.\\
Augmented: I have 2 pencils weighing 220 gms. I went out and bought 3 more. Now my pencils weigh 300 gms.\\
\textit{\textbf{Passage}: <input MWP>}  \\
\textit{\textbf{New Variable}: from Prompt 1} \\ 
\textit{\textbf{Start value}: from Prompt 1} \\
\textit{\textbf{End value}: from Prompt 1}

\section{Extended Related Work}
\cite{jayarao2021retrainingdistilbertvoiceshopping} showcase the ability of using encoder based SLMs to generate embeddings for task-oriented multi-turn dialogue systems. While \cite{9001577} showcases the impact of multilingual and code-mix training on language models.
\cite{Dwivedi_2022_CVPR} demonstrate the value of using disparate data sources during model training. \cite{pmlr-v161-mallick21a} apply Bayesian inference to improve model performance in small data regime. 
Recent work benchmarks Multimodal Large Language Model (MLLM) weaknesses in counting (`CountQA`~\cite{tamarapalli2025countqamllmscountwild}), perception (`HueManity`~\cite{grover2025huemanityprobingfinegrainedvisual}), and geographic reasoning (`GeoChain`~\cite{yerramilli2025geochainmultimodalchainofthoughtgeographic}). To address underlying flaws like unimodal dominance, research has explored multimodal attribution (`MAEA`~\cite{jain2023maeamultimodalattributionembodied}) and attribution regularization~\cite{yerramilli2025attributionregularizationmultimodalparadigms}, with direct applications in areas like assistive egocentric navigation~\cite{Jadhav_Cao_Shetty_Kumar_Sharma_Sukboontip_Tamarapalli_Zhang_Koul_2025}. Data-centric methods, such as semantic image augmentation using language, also offer a path to improve model robustness~\cite{yerramilli2024semanticaugmentationimagesusing}.

\end{document}